\documentclass[letterpaper, 10 pt, conference]{ieeeconf}

\IEEEoverridecommandlockouts
\usepackage{graphics}
\usepackage{epsfig}
\usepackage{times}
\usepackage{amsmath}
\usepackage{amssymb}
\usepackage[utf8]{inputenc}
\usepackage[hidelinks]{hyperref}

\usepackage[utf8]{inputenc} 
\usepackage[T1]{fontenc}    
\usepackage{url}            
\usepackage{booktabs}       
\usepackage{amsfonts}       
\usepackage{nicefrac}       
\usepackage{microtype}      
\usepackage{xcolor}         

\usepackage{graphicx}
\usepackage{caption}
\usepackage{booktabs}
\usepackage{amsmath}
\usepackage{algorithm}
\usepackage{algpseudocode}
\usepackage{amssymb}
\usepackage{multirow}
\usepackage{tabularx}
\usepackage{colortbl}  %
\usepackage{array}
\usepackage{siunitx} %
\usepackage{hyperref}
\usepackage{tikz}
\usepackage{tcolorbox}  %
\usepackage{listings}
\usepackage{mathrsfs}
\usepackage{wrapfig}
\usepackage{upgreek}
\usepackage{makecell}
\usepackage{vcell}
\usepackage{mwe} %
\usepackage{calc} %
\usepackage{pifont} %
\usepackage{marvosym}   %
\usepackage{wrapfig}
\usepackage{lipsum}
\usepackage[table]{xcolor}
\usepackage{xcolor}
\usepackage[capitalize]{cleveref}
\usepackage{acronym}
\usepackage[algo2e]{algorithm2e}
\usepackage[compact]{titlesec}
\usepackage{stfloats}      
\crefname{section}{Sec.}{Secs.}
\Crefname{section}{Section}{Sections}
\crefname{table}{Tab.}{Tabs.}
\Crefname{table}{Table}{Tables}
\crefname{figure}{Fig.}{Fig.}
\Crefname{figure}{Figure}{Figure}
\definecolor{revision}{RGB}{0,0,255}

\definecolor{iccvblue}{rgb}{0.21,0.49,0.74}

\definecolor{Gray}{gray}{0.85}
\newcommand{\sota}{state-of-the-art\xspace}
\newcommand{\method}{\mbox{{VLA-Reasoner}}\xspace}
\newcommand{\tech}{MCTS\xspace}

\definecolor{best}{rgb}{0.96, 0.57, 0.58}
\definecolor{second}{rgb}{0.98, 0.78, 0.57}
\definecolor{third}{rgb}{1.0, 1.0, 0.56}

\definecolor{shallowgreen}{RGB}{200, 230, 201} 

\definecolor{white}{RGB}{255,255,255}
\definecolor{ourcolor}{RGB}{251,234,255}
\colorlet{reasoner}{ourcolor!40!white}

\acrodef{vae}[VAE]{Variational Autoencoder}
\acrodef{il}[IL]{Imitation Learning}
\acrodef{rl}[RL]{Reinforcement Learning}
\acrodef{mbrl}[MBRL]{Model-based Reinforcement Learning}
\acrodef{pomdp}[POMDP]{Partially Observable Markov Decision Process}
\acrodef{mdp}[MDP]{Markov Decision Process}
\acrodef{3d}[3D]{three-dimensional}
\acrodef{nerf}[NeRF]{Neural Radiance Field}
\acrodef{sde}[SDE]{stochastic differential equation}
\acrodef{3dgs}[3D-GS]{3D Gaussian Splatting}
\acrodef{fps}[FPS]{Farthest Point Sampling}

\newlength\savewidth

\makeatletter
\renewcommand{\paragraph}{%
  \@startsection{paragraph}{4}{\z@}%
  {1ex plus 0.5ex minus 0.2ex} 
  {-1em}                      
  {\normalfont\normalsize\bfseries} 
}
\makeatother

\usepackage{amsmath,amsfonts,bm}

\def\eqref#1{equation~\ref{#1}}

\def\1{\bm{1}}

\DeclareMathAlphabet{\mathsfit}{\encodingdefault}{\sfdefault}{m}{sl}
\SetMathAlphabet{\mathsfit}{bold}{\encodingdefault}{\sfdefault}{bx}{n}

\usepackage{fancyhdr}
\makeatletter
\let\NAT@parse\undefined
\makeatother

\title{\LARGE \bf
VLA-Reasoner: Empowering Vision-Language-Action Models with Reasoning via Online Monte Carlo Tree Search
}

\author{
\textbf{Wenkai Guo$^{*,1}$},
\textbf{Guanxing Lu$^{*,2}$},
\textbf{Haoyuan Deng$^{1}$},
\textbf{Zhenyu Wu$^{3}$},
\textbf{Yansong Tang$^{2}$},
\textbf{Ziwei Wang$^{\dagger,1}$} \\
$^{1}$ School of Electrical and Electronic Engineering, Nanyang Technological University \\
$^{2}$ Tsinghua Shenzhen International Graduate School, Tsinghua University \\
$^{3}$ School of Intelligent Engineering and Automation, Beijing University of Posts and Telecommunications \\
\texttt{wenkai001@e.ntu.edu.sg, ziwei.wang@ntu.edu.sg}
}

\begin{document}
\maketitle
\thispagestyle{fancy}
\pagestyle{fancy}

\begin{abstract}
    Vision-Language-Action models (VLAs) achieve strong performance in general robotic manipulation tasks by scaling imitation learning. However, existing VLAs are limited to predicting short-sighted next-action, which struggle with long-horizon trajectory tasks due to incremental deviations.
    To address this problem, we propose a plug-in framework named \method that effectively empowers off-the-shelf VLAs with the capability of foreseeing future states via test-time scaling.
    Specifically, \method samples and rolls out possible action trajectories where involved actions are rationales to generate future states via a world model, which enables \method to foresee and reason potential outcomes and search for the optimal actions.
    We further leverage Monte Carlo Tree Search (MCTS) to improve search efficiency in large action spaces, where step-wise VLA predictions seed the root.
    Meanwhile, we introduce a confidence sampling mechanism based on Kernel Density Estimation (KDE), to enable efficient exploration in MCTS without redundant VLA queries.
    We evaluate intermediate states in MCTS via an offline value estimation strategy, to score predicted futures and correct deviations with long-term feedback.
    We conducted extensive experiments in both simulators and the real world, demonstrating that our proposed VLA-Reasoner achieves significant improvements over the state-of-the-art VLAs. Our method highlights a potential pathway toward scalable test-time computation of robotic manipulation. The project website is available at: \url{https://vla-reasoner.github.io/}.
\end{abstract}

\section{Introduction}
\label{sec:intro}

Vision-Language-Action models (VLAs) \cite{kim2024openvla,black2024pi0visionlanguageactionflowmodel,team2024octo} leverage the grounded perception and commonsense reasoning of large, pre-trained vision–language models (VLMs) to advance general-purpose robot manipulation. Within a supervised imitation learning paradigm, they map visual observations and natural-language instructions directly to sequences of low-level actions using extensive robot demonstration datasets \cite{collaborationOpenXEmbodimentRobotic2024a,rt12022arxiv,walke2023bridgedata}. By decoupling task specification from policy learning, VLAs adapt to diverse manipulation scenarios via language prompts and image conditions. Recent results show stronger generalization across object categories and environments, and reduced dependence on explicit task engineering, positioning VLAs as a promising route toward scalable embodied intelligence \cite{black2024pi0visionlanguageactionflowmodel,intelligence2025pi05visionlanguageactionmodelopenworld}.

\begin{figure}[t]
    \centering
    \includegraphics[width=0.48\textwidth]{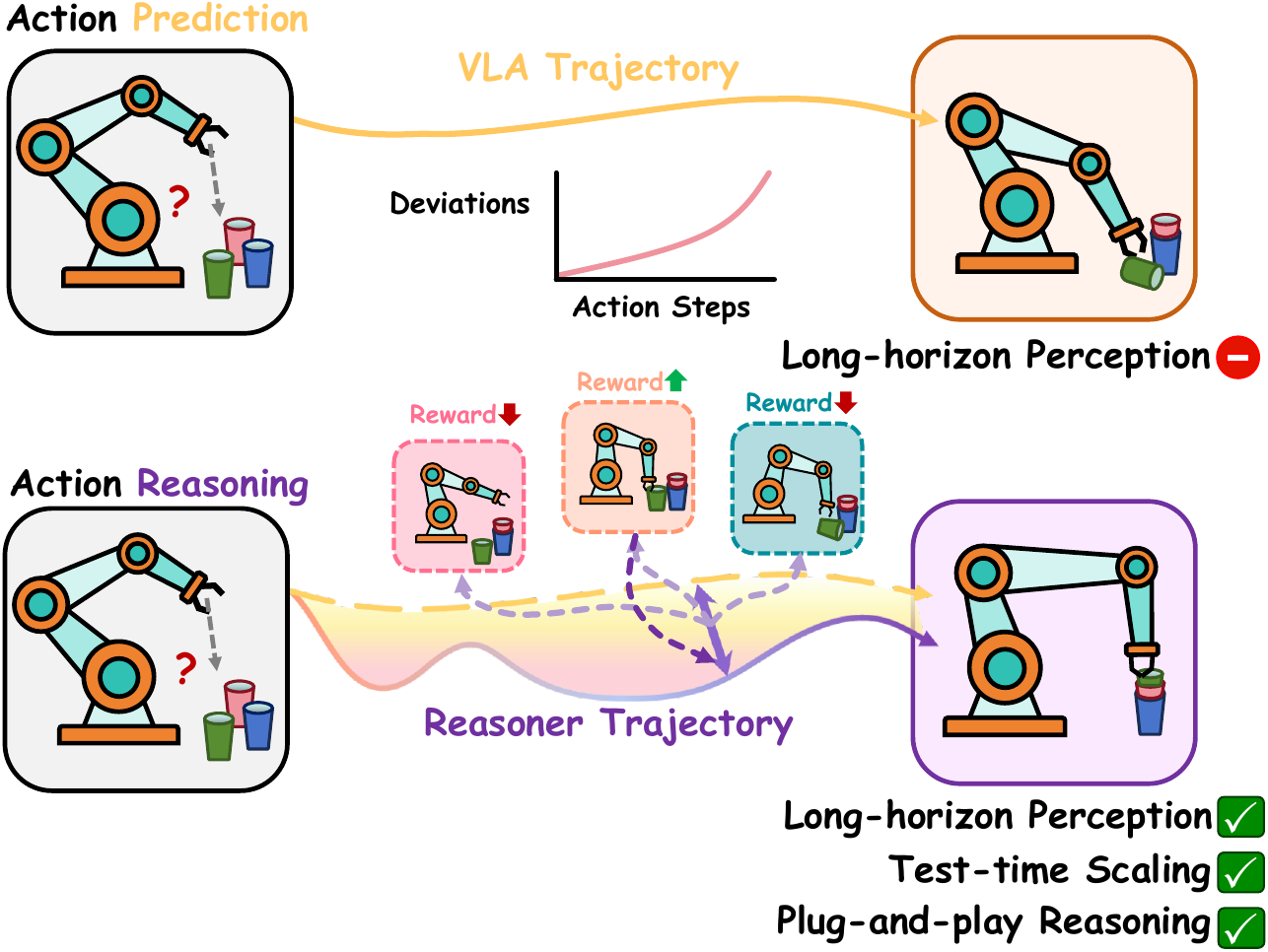}
    \caption{\small \textbf{VLA-Reasoner} augments VLA models with test-time reasoning via online tree search, enabling more robust and interpretable robotic manipulation than baselines.}
    \vspace{-0.1cm}
    \label{fig:teaser}
\end{figure}

However, current VLAs also face critical limitations. As the action prediction of VLAs fundamentally relies on direct mappings from short-horizon environment states to actions, they remain fragile during deployment. This short-sighted prediction discards long-horizon sequential dependencies, becoming a primary cause of incremental deviations across tasks and environments. Consequently, the accuracy and exploration capability of VLAs are significantly constrained. 
This raises a core question: \emph{``Can VLAs explore the long-horizon future influence of actions at test time, and decide the optimal action?''} 

To this end, we propose a plug-in framework named \method that empowers off-the-shelf VLAs with the ability to foresee future states via test-time scaling (\Cref{fig:teaser}).
Specifically, \method samples and rolls out possible action trajectories to generate future states via a world model, where the future states and corresponding actions can reflect the potential outcomes.  
To enhance search efficiency, we employ MCTS to handle the expansive action space, in which step-wise VLA predictions seed the root node.
We introduce a KDE-based confidence distribution that samples candidates in MCTS from an expert-like prior, reducing redundant VLA queries while preserving exploration.
Since sparse task feedback arrives only at episode ending, we design an offline value estimation strategy to evaluate intermediate states in MCTS, providing dense feedback signals that correct deviations with stable long-horizon guidance.
\method effectively improves the reasoning capability of VLAs in long-horizon trajectory tasks through enabling structured exploration in expansive action spaces and foreseeing the potential outcomes of the current action.

\begin{figure*}[t]
    \centering
    \includegraphics[width=\textwidth]{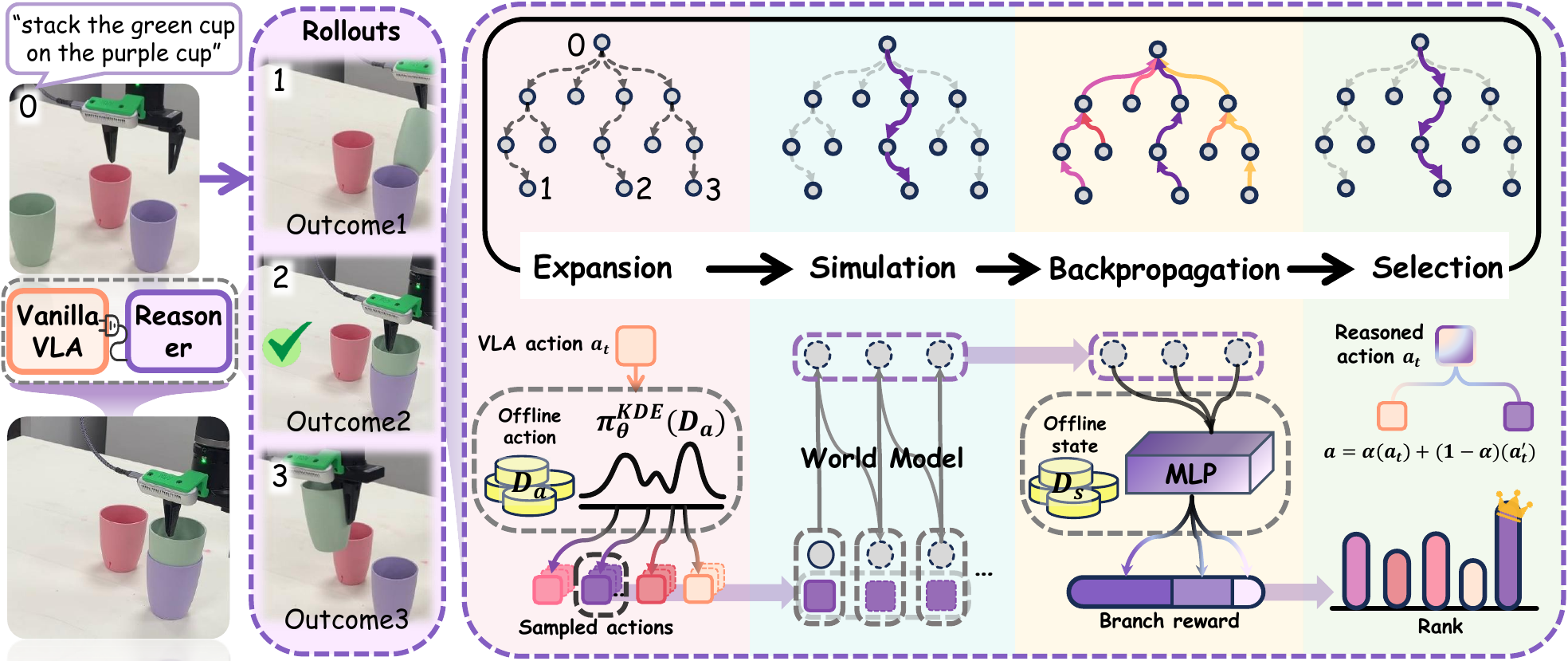}
    \caption{\small \textbf{The overall pipeline of VLA-Reasoner.} At test time, a lightweight and modified \tech searches for the optimal action conditioned on the VLA prediction. The search is steered by expert-like sampling and dense value estimation, which guide expansion and backup throughout the tree. The method is plug-and-play, and it can be attached to any VLA-based manipulation policy and consistently improves performance across tasks, environments, and robot embodiments.}
    \vspace{-0.5cm}
    \label{fig:pipeline}
\end{figure*}

Our method delivers consistent gains in both simulation and on real robots. On the LIBERO benchmark, wrapping a modest baseline VLA with \method lifts it beyond competing VLAs. In real-world deployments, our approach achieves higher success rates compared to popular VLAs fine-tuned with a few demonstrations, indicating stronger generalization and adaptivity at test time. Our contributions are summarized as follows:

\begin{itemize}
    \item We propose a plug-in framework named \method that empowers VLAs with structured reasoning to address their incremental deviations during deployment.

    \item  We adapt a modified test-time \tech to search efficiently rather than just use it. We apply KDE for efficient plausible expansion, and provide an offline-based value estimation method to evaluate intermediate states.
    
    \item We conduct extensive experiments that validate the effectiveness of integrating MCTS into VLAs for test-time optimization. We also show the potential to achieve great real-world performance with a few data acquisitions.
\end{itemize}

\section{Related Work}
\label{sec:related_work}

\noindent \textbf{Vision-Language-Action Models.} As robot learning increasingly demands generalist policies, Vision-Language-Action (VLA) models have emerged as a promising paradigm \cite{kim2024openvla,rt22023arxiv,black2024pi0visionlanguageactionflowmodel,team2024octo}. Recent efforts integrate Vision-Language Models (VLMs) into VLA systems, leveraging their broad vision-language understanding and reasoning capabilities \cite{touvron2023llama,wang2024qwen2}. Trained on large-scale robotic datasets \cite{collaborationOpenXEmbodimentRobotic2024a,walke2023bridgedata}, VLA models achieve state-of-the-art manipulation performance. Their generalization, inherited from VLMs, enables rapid adaptation to novel and unstructured environments \cite{li2023vision,stone2023open}.
Despite this progress, current VLAs still face challenges. Reliance on expert data and limited use of VLM reasoning restricts the generalization and transferability of VLA policies across domains. Recent work improves VLAs via low-level designs like data augmentation, action re-ranking, and reinforced fine-tuning \cite{xue2025demogen,nakamoto2024steering,lu2025vla}, but these gains are often incremental and brittle to distribution shifts in deployment. 
To mitigate the compounding errors in deployment of VLAs, \method introduces explicit test-time reasoning through model-guided tree search, systematically enhancing both the reasoning ability and adaptability of VLA deployment.

\noindent \textbf{Monte Carlo Tree Search.} Monte-Carlo Tree Search (MCTS) is a popular Markov Decision Process (MDP) planning algorithm to explore complex action decision space \cite{swiechowski2023monte,pitanov2023monte}. MCTS especially serves as a reasoning augment mechanism in deep learning from agent learning in game decision \cite{schrittwieser2020mastering, perez2014knowledge} and Large Language Model (LLM) optimization \cite{yao2024mulberry,snell2024scaling}. The Tree Search algorithm performs strong self-guided reasoning ability to solve long and difficult problems, therefore, it plays a heuristic role in our method. MCTS has been adapted in robot control and planning \cite{vagadia2024phyplan, DecMCTSDecentralizedPlanning, leisiazar2023mcts}. Prior work mainly targets path planning on discrete graphs (e.g., go), which differs fundamentally from model-driven robotic manipulation with VLAs. In deployment, VLAs should reason over multi-modal observations and contact-rich, non-stationary dynamics, which makes vanilla MCTS difficult to run online. We therefore adapt MCTS for robotic manipulation by leveraging efficient simulated interactions, rather than using it directly.

\noindent \textbf{Model-based Planning.}
Classical model-based planning simulates a dynamics model forward and selects the action sequence with the lowest cost (or highest reward). Trajectory-optimization and MPC variants have been widely used in robotics with analytic or calibrated models \cite{tassa2012synthesis,williams2017information}.
With learned dynamics, ``world models'' replace physics engines and enable planning directly from images—e.g., latent-space video models and visual-foresight methods that predict future frames for goal-conditioned manipulation \cite{ha2018RecurrentWorldModels,hafner2019dream}. More recently, sequence models plan by proposing and scoring full trajectories \cite{janner2021trust,janner2022planning}. Our setting departs from these in two key aspects. First, we attach planning at test time to a pretrained VLA: the VLA's one-step prediction serves as the root, and a learned world model is used solely for online look-ahead rather than policy training. Second, instead of continuous trajectory optimization, we adopt discrete tree search guided by a data-driven action prior and dense visual value estimates, thereby avoiding hand-tuned costs and task-specific heuristics.

\section{Method}
\label{sec:method}

In this section, we first show the pipeline of our framework as ~\Cref{fig:pipeline}, and then present the formulation of our work (\Cref{subsec:prelimilaries}). We adapt \tech (\Cref{subsec:mcts}) for efficient test-time expansion and backpropagation on the VLA prediction without disturbing real world execution. Under MCTS manner, we apply KDE to efficiently sample action candidates for exploration rather than repeatedly querying VLA (\Cref{subsec:KDE}). We also estimate state values based on offline data to densely evaluate intermediate candidate-driven states (\Cref{subsec:value}).

\subsection{Problem Statement}\label{subsec:prelimilaries}
VLAs aim to generalize robot manipulation by mapping multimodal inputs (states from the environment $s_t$, language instructions of the task $l$) to actions $a_t^{VLA}$. The prediction of pretrained VLA can be formulated as $a_t^{VLA} = \pi_\theta(s_t, l)$ where $\theta$
means the parameters of the model and is frozen. Since the prediction relies only on the current state $s_t$, the absence of future consideration leads to deviations that progressively grow over time. 

Our goal is to mitigate this deviation during test time, which enables foreseeing the future impact of current action.
Instead of directly deploying VLA, we simulate future states using a world model $\mathcal{W}$ and score them with a value estimate $v$. At timestamp $t$, we feed action $a_t$ and state $s_t$ into $\mathcal{W}$ and get feedback of next state $s_{t+1}$, this process can be represented as $\mathcal{W}(s_{t+1} \mid a_t, s_t)$. After simulations under a MCTS manner, \method rolls out an action  $a_t^{Reasoner}$. We then inject $a_t$ with \method:
\begin{equation}
a_t = \alpha(a_t^\text{VLA})+(1-\alpha)(a_t^\text{Reasoner})
\label{eq:reasoner}
\end{equation}
where $\alpha$ represents the injection strength to control the balance of two actions. The $a_t^\text{Reasoner}$ integrates future impact into the decision, providing long-horizon guidance, whereas the final action $a_t$ represents the executed decision during deployment.

\subsection{Online Monte Carlo Tree Search}\label{subsec:mcts}

The key to \method lies in leveraging a tree structure consist of possible action trajectories and corresponding states for guided and directional search, where reasoning is enabled with interaction in a world model. We follow the MCTS manner for efficient tree search, and adapt MCTS to a simple implementation in test time. At each step $t$, \method proceeds through four steps:
\textbf{(a) Expansion}, \textbf{(b) Simulation}, \textbf{(c) Backpropagation}, and \textbf{(d) Selection}. A node in the MCTS process is indexed by $i$, denoted as $o_i$, with its associated state $s_i$ and value estimate $v_i$. The corresponding action to transit into this node is $a_i$. We further introduce the specific implementation of each step as follows.

\paragraph{Expansion}\label{par:expansion} The expansion step aims to expand the selected node $o_i$ (initially the root node, the superscripts from here to \hyperref[par:selection]{Selection}
 are slightly obfuscated to make them intuitive), to generate its children nodes. As actions are directly related to the generation of new states, we sample a set of actions $ \mathcal{A}= \{a_1, a_2, \ldots, a_k\}$ from a distribution $\pi_{\theta}$ under a Top-k strategy. For action $a_i$, the expansion can be formulated as:

\begin{equation}
\label{eq:sample}
\begin{aligned}
{\textbf{Sample: }}\;
\tilde{\mathcal{A}}_i &= \{ a^{(n)} \}_{n=1}^{N}\sim \pi_{\theta},\\[2pt]
{\textbf{Top-$k$: }}\;
\mathcal{A}_i^{\text{Top-k}}
&= \operatorname*{arg\,min}_{A \subseteq \tilde{\mathcal{A}}_i,\, |A|=k}
\;\sum_{a \in A} \| a - a_i \|_2
\end{aligned}
\end{equation}
where we get the $\mathcal{A}_i^{\text{Top-k}}$ as the candidates to expand from a randomly large sample set $\widetilde{\mathcal{A}}_i$, which are closest to the $a_i$ in Euclidean distance ($k$ here is relatively small compared to $N$).

\paragraph{Simulation}\label{par:simulation} To explicitly evaluate the influence of these sampled actions, we simulate future states using a learned action-aware world model~\cite{wu2024ivideogpt} as the backbone to predict visual states conditioned on actions. The simulation formulates:

\begin{equation}
s_{i+1} =\mathcal{W}(a_i, s_i)
\label{eq:simulate}
\end{equation}
where the world model rolls out the next state $s_{i+1}$ under a given action $a_i$ and current state $s_i$. We embed robot actions into its latent space and finetune it on a small robot dataset, aligning multimodal inputs for plausible transition generation.

\paragraph{Backpropagation}\label{par:backpropagation} As the tree search meets the truncating condition, \tech updates the node value in bottom-up order, from each leaf node to the root. The overall value of node $o_i$ is $\mathcal{Q}(o_i)$, which is balanced by combining the cost of the visit count $N(a_i)$. The value and visit count are updated:

\begin{equation} 
\begin{aligned} 
N(o_i) & = \sum_{a \in \mathcal{A}_i^{\text{Top-k}}} N(a), \\ \mathcal{Q}(o_i) & = \frac{N(o_i)\, v_i + \sum_{j \in \text{children}} N(o_j)\, \mathcal{Q}(o_j)} {N(o_i) + \sum_{j \in \text{children}} N(o_j)}
\end{aligned} 
\label{eq:update} 
\end{equation}
where $v_i$ is the value estimate of node $o_i$ and $children$ are the candidates of expansion. Since visit counts dominate inference cost, we estimate the visit count of a sampled action by its probability density naturally derived from the distribution (\Cref{subsec:KDE}). The value estimation is introduced in \Cref{subsec:value}.

\paragraph{Selection}\label{par:selection} This step aims to select the preferred node that balances search quality and efficiency. The selection follows two factors: the value $\mathcal{Q}(o_i)$ and the visiting counts $N(o_i)$ of the node $o_i$. We adopt the Upper Confidence Bound (UCB) strategy \cite{kocsis2006bandit}. The selection follows the formula:
\begin{equation}
a_{selected} =  \operatorname*{arg\,max}_{i \in k} \mathcal{Q}(o_{i}) + c \cdot \sqrt{\frac{\ln N(\hat{o_i})}{1 + N(o_i)}} 
\label{eq:select}
\end{equation}
where $c$ is a constant to constrain the exploration (e.g., $\frac{1}{\sqrt{2}}$),  $\hat{o_i}$ stands for the parent node of $o_i$. The UCB strategy enables a balance between exploration and exploitation during search. The node is selected with the highest UCB score.

These four steps are repeated in a round of iteration, where it takes real state and action as input. The whole process constructs an independent Monte Carlo Tree of current robot states as we use a world model to dictate the transitions. Generating such a tree structure represents finding an optimized search space of actions, in which we can rollout the best candidate. Then inject this candidate into the VLA prediction following \Cref{eq:reasoner}.
As VLA can always predict next-best action, we sparsely conduct \tech on VLA generations, where the efficiency can be apparently optimized while action quality is improved.

\subsection{Distribution for Efficient Sampling}\label{subsec:KDE}
To efficiently sample actions for expansion, we utilize Kernel Density Estimation (KDE) to model the distribution of actions from offline data. KDE is a non-parametric way to estimate the probability density function of a random variable, which allows us to generate diverse action candidates that are likely to be effective based on historical data. With a dataset of actions $\{a_1, a_2, \ldots, a_n\}$, the KDE can be formulated as:
\begin{equation}
\begin{aligned}
\pi_{\theta}^{\text{KDE}}(a) &= \frac{1}{N}\sum_{i=1}^{N} K_h(a-a_i),\\
\end{aligned}
\label{eq:kde}
\end{equation}
where $K$ is the kernel function (we use a Gaussian kernel here), and $h$ is the bandwidth parameter that controls the smoothness of the estimated density, $\theta$ means the above involved hyperparameters. 

We can then efficiently sample the distribution via $a_i \sim \pi_{\theta}^{\text{KDE}}(\cdot)$.
Since KDE returns a probability density $p_i= \pi_{\theta}^{\text{KDE}}(a_i)$, we can treat it as a Monte-Carlo estimate of how often the action (and its corresponding state) would be visited under large-scale sampling. In practice, this gives a soft prior for the visit count as $N(a) \propto p(a)$, which is crucial for efficient backpropagation.

For policies that output action chunks (e.g., a sequence of actions), we adapt the KDE to sample action chunks by treating each chunk as a single entity. To avoid long chunks washing out fine-grained corrections, we modify the used policy to roll out in a controllable size of actions (e.g., below 8) while still benefiting from sequence-level proposals.

\subsection{Vision-based Value Estimation}\label{subsec:value}
Accurate value estimation of intermediate states is essential for effective tree search. To evaluate the value of model-generated states, we follow the idea that changes in visual observations are a key indicator of task progress \cite{ma2024vision}. However, manipulation trajectories are usually dense. To reduce redundancy and preserve meaningful changes, we down-sample image sequences from the offline dataset, ensuring that consecutive frames contain sufficient variation while retaining most progress information. Based on the down-sampled sequence, we assign ground-truth value labels through linear interpolation between 0 and 1. For example, in a 10-frame sequence, the $5^{th}$ frame is assigned a value of $\tfrac{5}{9}$.

For efficient and consistent scoring, we utilize the ResNet-34 \cite{he2015deepresiduallearningimage} with frozen ImageNet-pretrained weights as the visual encoding backbone, followed by a 2-layer MLP training with MSE loss. The training objective can be formulated as:

\begin{equation}
\begin{aligned}
& \psi^\star = \operatorname*{arg\,min}_{\psi}\,
  \mathcal{L}_{\mathrm{MSE}}\!\left(\mathrm{MLP}(s_{t}), \{v_{t}\}\right) \\[3pt]
\end{aligned}
\label{eq:value}
\end{equation}
where the value is estimated with $v_{t} = \mathrm{MLP}(s_{t})$. With a simple but efficient design (the network takes less than 30 minutes to train), our design enables accurate value estimation to boost the performance (\Cref{subsec:ablation}).

Pseudocode for \method is provided above (\Cref{alg:reasoner}). Components except the world model are trained on the same dataset used to finetune the VLA. For the world model, we additionally collect a small set of failure demonstrations to finetune it for predicting failure cases.

\begin{figure}[t]
\centering
\begin{minipage}{0.44\textwidth}  
\vspace{-0.2cm}
\begin{algorithm}[H]
\small
\caption{VLA-Reasoner}
\label{alg:reasoner}
\DontPrintSemicolon
\SetAlgoLined
\LinesNumbered
\SetKwInOut{Prep}{Preparation}
\SetKwInOut{Input}{Input}
\SetKwInOut{Output}{Output}
\SetKw{KwTo}{to}
\SetKwFunction{SelectUCB}{SelectUCB}
\SetKwFunction{TopK}{TopK}
\SetKwFunction{Update}{Update}
\SetKwComment{tcp}{\footnotesize\textit{// }}{}

\Prep{%
world model $\mathcal{W}$, KDE prior $\pi_{\theta}^{\mathrm{KDE}}$, value network $\mathrm{MLP}$.
}

\Input{VLA proposal $a_t^{\text{VLA}}$, current state $s_t$}
\Output{final action $a_t$}

\textbf{Init:} Create root node $o^{(0)}$ with $s^{(0)} \gets s_t,\, a^{(0)} \gets a_t^{\text{VLA}}$.\,;

\For{depth $d=0$ \KwTo $MaxDepth$}{

  \tcp{Expansion}
  $\mathcal{A}^{\text{Top-k}} \gets$ \TopK{$\pi_{\theta}^{\mathrm{KDE}}, a^{d}$}   \Comment{\textit{\Cref{eq:sample}}}

  \For{$a \in \mathcal{A}^{\text{Top-k}}$}{
    \tcp{Simulation}
    $\hat{s} \gets \mathcal{W}(s^{(d)}, a)$\;

    \tcp{Evaluation}
    $\hat{v} \gets \mathrm{MLP}(\hat{s})$ 

    \tcp{Backpropagation}
    \Update{$o^{(d)},\,\hat{v}$}  \Comment{\textit{\Cref{eq:update}}}\;
  }
  \tcp{Selection}

  $o^{(d)} \gets$ \SelectUCB{$o^{(d)}$} \Comment{\textit{\Cref{eq:select}}} 
  
  $a^{(d)} \gets a(o^{(d)})$

}

\tcp{Reasoner rollout}
$a_t^{\text{Reasoner}} \gets \operatorname*{arg\,max}_{a_i \in \text{children}} \mathcal{Q}(o_i)$\;

\tcp{Action injection with strength $\alpha$}
$a_t \gets \alpha \cdot a_t^{\text{VLA}} + (1-\alpha) \cdot a_t^{\text{Reasoner}}$\;

\end{algorithm}
\end{minipage}
\end{figure}

\section{Experiments}
\label{sec:experiments}

\begin{table*}[t]
\centering
\normalsize
\setlength{\tabcolsep}{5pt}
\renewcommand{\arraystretch}{1.15}

\caption{\small \textbf{Results in Simulations.} Average success rates across 500 episodes for LIBERO and 100 episodes for SimplerEnv. Our method outperforms OpenVLA-SFT on all 4 direction tasks and Octo-Small/SpatialVLA on 4 tasks. Bold entries mark the highest success rates, underlined for second-best. Asterisked results are chosen baselines and locally evaluated for fairness.}
\vspace{0pt}\centering
\begin{tabular}{@{}%
p{3.6cm}  
>{\centering\arraybackslash}p{2.2cm}
>{\centering\arraybackslash}p{2.2cm}
>{\centering\arraybackslash}p{2.2cm}
>{\centering\arraybackslash}p{2.2cm}
>{\centering\arraybackslash}p{2.2cm}
@{}}

\toprule
\textbf{LIBERO} & Spatial & Goal & Object & Long & Average \\
\midrule
Diffusion Policy \cite{chi2024diffusionpolicy} & 78.3\% & 68.3\% & \textbf{92.5\%} & 50.5\% & 72.6\% \\
GRAPE \cite{zhang2024grape} & 87.6\% & \underline{82.2\%} & \underline{91.2\%} & \underline{55.8\%} & \underline{79.2\%} \\
TraceVLA \cite{zheng2024tracevla} & 84.6\% & 75.1\% & 85.2\% & 54.1\% & 74.8\% \\
SpatialVLA \cite{qu2025spatialvla} & 88.2\% & 78.6\% & 89.9\% & 55.5\% & 78.1\% \\
\textsuperscript{*}OpenVLA-SFT \cite{kim2024openvla} & 84.6\% & 79.2\% & 86.6\% & 53.4\% & 76.0\% \\
\rowcolor{reasoner}
+\method & \textbf{91.2\%} & \textbf{82.4\%} & 90.6\% & \textbf{59.8\%} & \textbf{81.0\%} \\
\midrule
\rowcolor{white}
\textbf{SimplerEnv} & Block & Spoon & Carrot & Eggplant & Average\\
\midrule
\rowcolor{white}
\textsuperscript{*}Octo-Small\cite{team2024octo} & 5\% & 43\% & 10\% & 48\% & 26.5\%\\
\rowcolor{reasoner}
+\method & \textbf{13\%} & \textbf{50\%} & \textbf{32\%} & \textbf{54\%} & \textbf{37.3\%} \\
\rowcolor{white}
\textsuperscript{*}SpatialVLA & 23\% & 18\% & 22\% & 73\% & 34.0\%\\
\rowcolor{reasoner}
+\method & \textbf{29\%} & \textbf{35\%} & \textbf{28\%} & \textbf{75\%} & \textbf{41.8\%}\\
\bottomrule
\end{tabular}
\label{table:simulation-results}
\vspace{-0.4cm}
\end{table*}

In this section, we present both simulation and real-world experiments to explore the following key questions:

\begin{enumerate}
\item \textbf{Test-time gains.} How does \method enhance the performance of different pretrained VLAs across diverse environments during the test time?
\item \textbf{Real-world Applicability.} How does \method help generalist VLAs in real-world manipulation tasks?
\item \textbf{Robustness.} Can \method adapt to varied settings while maintaining performance?  
\end{enumerate}

For \textbf{Q1}, we conduct experiments in $2$ simulation environment (LIBERO~\cite{liu2023libero} and SimplerEnv~\cite{li2024evaluating}) with $8$ specific tasks based on $3$ popular general robot policies. 
For \textbf{Q2}, we evaluate the practicality of the \method on $5$ challenging real-world tasks on top of $2$ advanced VLAs.
For \textbf{Q3}, we ablate two main factors (one supporting our method and the other one supporting VLA deployment) to assess adaptation and stability. We also conduct ablation on specific technique designs to test the effectiveness.

\subsection{\method in Simulation}
\label{subsec:simulation}

\paragraph{Experiment Setup}\label{par:setup_sim} We evaluate our method with a combination of $3$ popular generalist robot policies: OpenVLA, which is a VLA model with a parameter size of 7B \cite{kim2024openvla}; Octo-Small \cite{team2024octo}, which is a transformer combining a diffusion head for action prediction, with a parameter size of 27M; SpatialVLA, which is a spatial-enhanced VLA model with a parameter size of 4B \cite{qu2025spatialvla}. We follow the architecture of iVideoGPT \cite{wu2024ivideogpt} to train an action-aware world model, with a parameter size of 600M. All the training phases (including KDE estimation and value estimation) rely on the same public datasets generated from the simulators. For the world model, we additionally supplement its training with a small set of failure demonstrations collected from the rollouts of the pretrained VLA itself, enabling the model to capture those failures during deployment. For execution, OpenVLA and Octo roll $1$ action every time, and SpatialVLA rolls $4$ actions as a chunk every time. We choose LIBERO \cite{liu2023libero} and SimplerEnv \cite{li2024evaluating} for simulation. 
For LIBERO, we utilize $4$ task suites: Spatial, Object, Goal, and Long. Each suite contains $500$ expert demonstrations distributed across $10$ language-conditioned tasks with $50$ variations, designed to evaluate policy generalization across different spatial configurations, object types, goals, and long-horizon task sequences. In SimplerEnv, we employ $4$ representative tasks on the WidowX robot: 1. \textbf{Block}: \emph{Stack green block on yellow block}; 2. \textbf{Spoon}: \emph{Put spoon on towel}; 3. \textbf{Carrot}: \emph{Put carrot on plate}; 4. \textbf{Eggplant}: \emph{Put eggplant in yellow basket}. We use OpenVLA-SFT to refer to the OpenVLA model finetuned on the public LIBERO dataset. All the training processes are conducted on a server with 6 NVIDIA RTX 6000 GPUs.

\paragraph{Quantitative Study}\label{par:quantitative} Experiment results are shown in \Cref{table:simulation-results}. As the success rate is the primary metric of evaluation in two benchmarks, our method improves the absolute task-set performance on OpenVLA-SFT by 5\% on average, the reasoner also improves the task performance on Octo-Small by 9.8\%, and SpatialVLA by 7.8\%, on average. 
It is noticeable that compared to those variants developed from OpenVLA, our plug-and-play method can directly improve the performance of the backbone to the \sota level without large-scale and skillful post-training, which is required for those variants. When compared to SpatialVLA, which is augmented with better spatial understanding capability, our method outperforms it in all task suites. 
Moreover, our method shows notable improvements on inherently difficult tasks such as stack block, which typically require precise sequential reasoning and are sensitive to slight prediction deviations. By introducing step-wise simulated inference, \method enables more deliberate future planning, reducing incremental deviations and significantly boosting task success.

\paragraph{Qualitative Study}\label{par:qualitative} The gains stem from test-time tree search that spawns counterfactual branches, exploring alternative futures without disturbing real execution. As our method captures future deviations caused by current actions, the backpropagation offers a look-ahead evaluation, bringing the Markovian deployment with a longer horizon to mitigate the limitation. To support efficient deployment, we finetuned the world model with the same data size as finetuning VLA, reducing computational cost and dataset reliance. Detailed analysis is provided in the ablation study.

\subsection{Deployment in Real-world Environment} \label{subsec:realworld}

\begin{figure}[t]
\centering
\includegraphics[width=\linewidth]{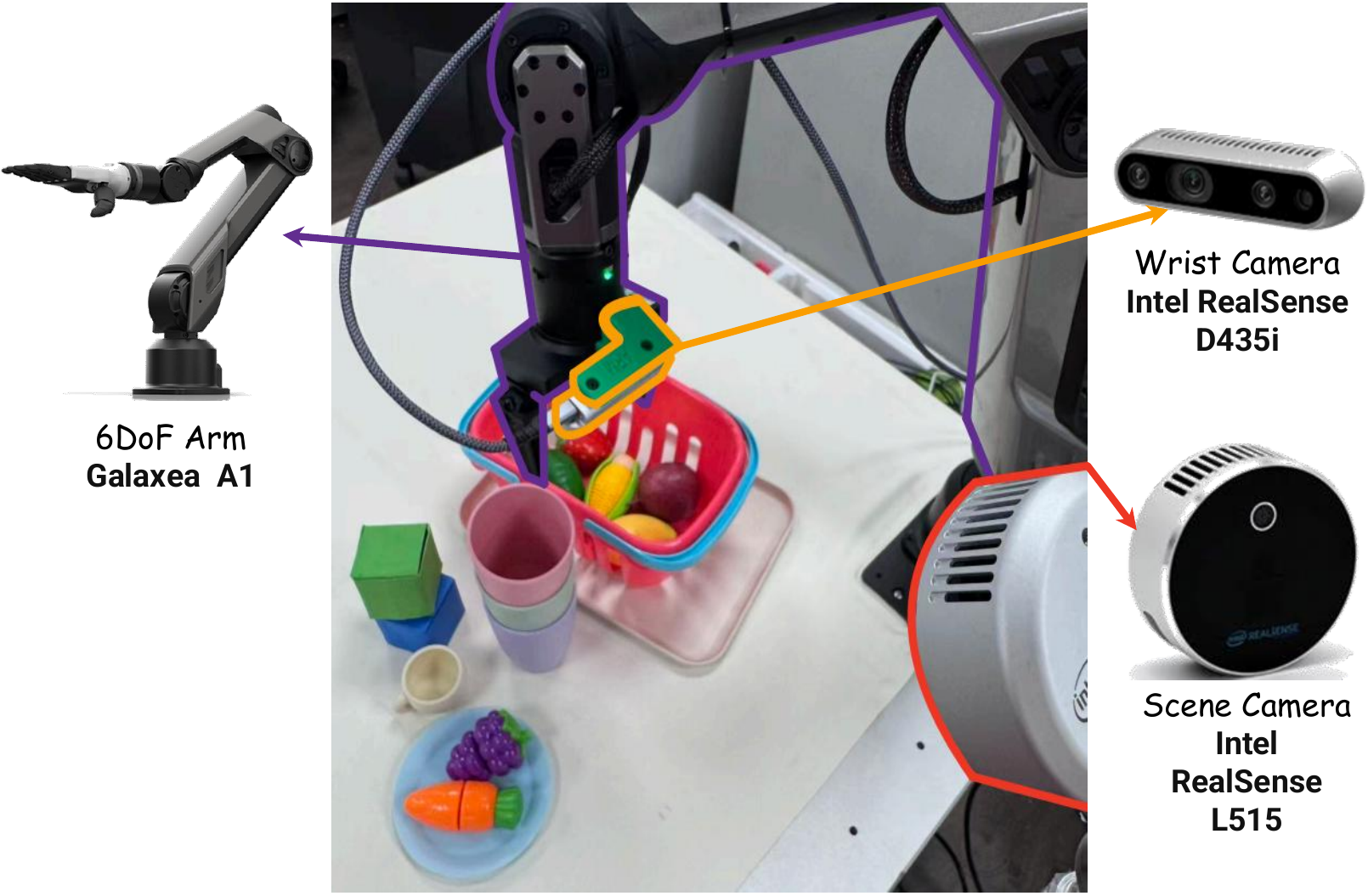}
\caption{\small \textbf{Setup of real world experiments.} We conduct diverse tasks in the real world to identify the limitations of current VLAs and validate our method.}
\label{fig:real_setup}
\end{figure}

\paragraph{Experiment Setup}\label{par:setup_real} To evaluate the performance of the \method in the real world with real robots. We design and test $5$ real-world tasks via deploying our method on two cutting-edge VLAs: an open-sourced popular model OpenVLA-7B \cite{kim2024openvla}, and an advanced commercial model $\pi_0-\text{FAST}$~\cite{pertsch2025fast} with local finetuning. The training phases use the same datasets, and we collect 10 failure cases for each task to supplement the training of the world model. After every inference, OpenVLA predicts the next $1$ action and $\pi_0-\text{FAST}$ predicts the next $5$ actions as a chunk. We evaluate with a fixed Galaxea-A1 robot arm. Images are provided by one side-fixed camera and one wrist-fixed camera (the arm-fixed camera is not used for OpenVLA since it does not support wrist image input). Real-world inference is conducted on an NVIDIA RTX 4090 GPU. The setup is visualized in ~\Cref{fig:real_setup}.

For each evaluation task, we collect 20 demonstrations with the initial positions of the primary objects slightly randomized. We design 5 tasks that test physical awareness, manipulation precision, and long-horizon understanding. The tasks include:

\begin{enumerate}
\item \textbf{Block}: \emph{Stack the green cube on the blue cube.}
\item \textbf{Fruit}: \emph{Pick the grape and place the grape into the basket.}
\item \textbf{1 Cup}: \emph{Stack the green cup on the purple cup.}  
\item \textbf{2 Cups}: \emph{Stack the green cup and the pink cup on the purple cup.}
\item \textbf{Circle}: \emph{Circle around the cup.}
\end{enumerate}

\paragraph{Results}\label{par:result_real}
 Experiment results are presented in \Cref{table:real-results}. In real-world experiments, our method can significantly improve finetuned VLA models during deployment. It averages an improvement of OpenVLA with an absolute gain of 19\%, a relative gain of 86.4\%, as the baseline shows a poor performance of 22\% success rate. It can also surprisingly boost $\pi_0\text{-FAST}$ with absolute gain of 10\%, relative gain of 15.6\%. Besides the strengths shown in \Cref{subsec:simulation}, we find that injecting a directional future-conditioned feedback to action can improve the awareness of current execution, and thus avoid failure modes. As VLAs struggle to succeed in the real world due to environment shifting and the embodiment gap, we analyze a specific case to reveal the short-sighted incremental deviations during real-world deployment, and showcase how our approach mitigates these kinds of deviations without disturbing action execution. 

\begin{figure*}[t]
    \centering
    \includegraphics[width=0.96\textwidth]{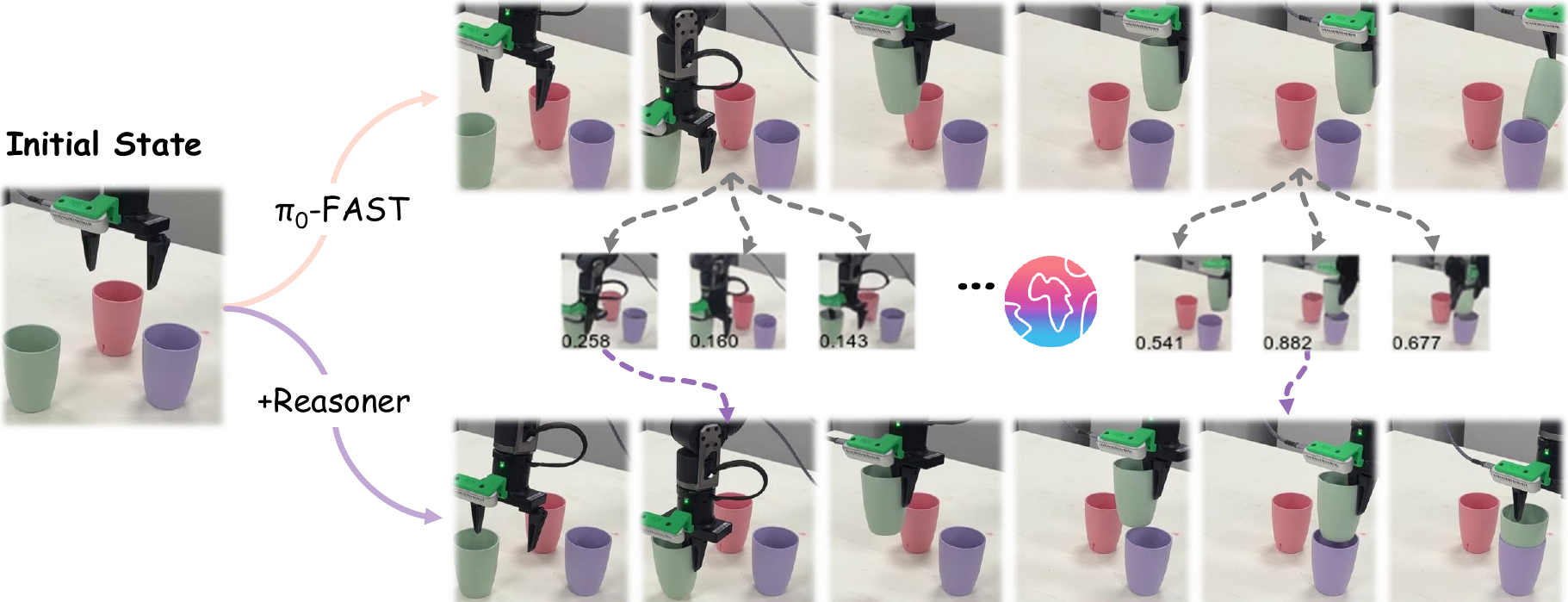}
    \caption{\small \textbf{Case Visualization.} The baseline policy ($\pi_{0}$-FAST, top row) suffers from excessive action drift and fails by such deviations. With reasoning, \method (bottom row) proactively corrects misalignment via value-guided search, enabling success.}
    \label{fig:case_study}
    \vspace{-0.4cm}
\end{figure*}

\paragraph{Case Study}\label{par:case}
Humans naturally reason over space and time when manipulating objects; the same foresight should be built into robotic policies. Without it, policies tend to produce unstable control and suboptimal decisions in fine-grained tasks. We examine a representative challenge: \emph{Stack the green cup on the purple cup}, which requires precise object localization, adaptive grasping, and fine-grained motion planning under visual distraction. As shown in \Cref{fig:case_study} (top), the baseline $\pi_0-\text{FAST}$ fails to properly align with the target due to the deviations from accumulative immature movement. In contrast, \method performs test-time simulation of future possible cases, enabling proactive correction before execution. As illustrated in \Cref{fig:case_study} (bottom), the reasoning phase conducted in the world model reflects the most possible future states via the value estimates (annotated on the lower left corner), and the action candidate with highest value is chosen to inject into the VLA rollout action, thus the robot adjusts its waypoint mid-trajectory and achieves a more stable, deliberate manipulation. This case reveals how \method empowers VLAs with reasoning ability during their deployment.

\begin{table}[t]
\centering
\normalsize
\setlength{\tabcolsep}{8pt}
\renewcommand{\arraystretch}{1.12}
\vspace{-0.1cm}
\caption{\small \textbf{Results in Real World.} Average success rates of 5 tasks in different scenarios. Each task is evaluated 20 times. Our method apparently improves OpenVLA and $\pi_0-\text{FAST}$ in all tasks.}
\begin{tabular}{l c c c}
\toprule
\textbf{Method} & \multicolumn{1}{c}{\makecell{\textbf{Block}}} 
& \multicolumn{1}{c}{\makecell{\textbf{Fruit}}} 
& \multicolumn{1}{c}{\makecell{\textbf{1 Cup}}} \\
\midrule
OpenVLA & 25\% & 45\% & 20\% \\
\rowcolor{reasoner}
+VLA-Reasoner & \textbf{40\%} & \textbf{70\%} & \textbf{40\%}\\
$\pi_0-\text{FAST}$ \cite{pertsch2025fast} & 70\% & 80\% & 60\% \\
\rowcolor{reasoner}
+VLA-Reasoner & \textbf{80\%} & \textbf{90\%} & \textbf{75\%} \\
\midrule
\textbf{Method} & \multicolumn{1}{c}{\makecell{\textbf{2 Cups}}} 
& \multicolumn{1}{c}{\makecell{\textbf{Circle}}} 
& \multicolumn{1}{c}{\makecell{\textbf{Overall}}} \\
\midrule
OpenVLA & 5\% & 15\% & 22\% \\
\rowcolor{reasoner}
+VLA-Reasoner & \textbf{15\%} & \textbf{40\%} & \textbf{41\%} \\
$\pi_0-\text{FAST}$ & 40\% & 70\% & 64\% \\
\rowcolor{reasoner}
+VLA-Reasoner & \textbf{50\%} & \textbf{75\%} & \textbf{74\%} \\
\bottomrule
\end{tabular}
\label{table:real-results}
\end{table}

\subsection{Ablation Analysis} \label{subsec:ablation}
This section aims to evaluate the robustness and sensitivity of \method under different injection strengths, and to validate whether its outstanding performance gains arise from our designs rather than other possible solutions. We conduct controlled ablations on LIBERO-Spatial. Two key factors are studied: (1) the \textit{Injection Strength} factor $\alpha$ to control the mix of VLA rollouts and sampled actions, and (2) the main techniques, including \textit{Sampling Criteria} and \textit{Value Estimation}. The base model remains identical to the main experiments.

\begin{figure}[t]
  \centering
  \includegraphics[width=\columnwidth]{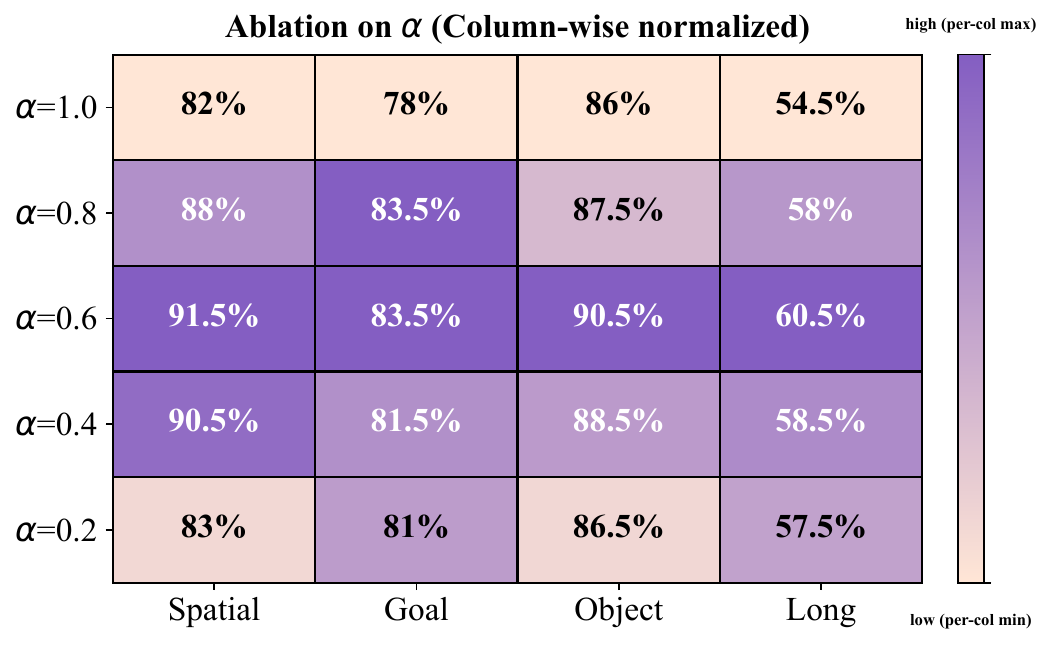}
  \caption{\small\textbf{Analysis on injection strength $\alpha$.} $\alpha$ controls the trade-off between the VLA action and the reasoner action; a larger $\alpha$ assigns greater weight to the VLA action. $\alpha=1.0$ means the vanilla VLA.}
  \label{fig:alpha_heatmap}
  \vspace{-0.4cm}
\end{figure}

\paragraph{Choices of Injection Strength}\label{par:inject}
~\Cref{fig:alpha_heatmap} shows that injecting a value-guided action by MCTS into VLA rollout boosts success performance under all $4$ settings.  $\alpha=0.6$ shows the highest success rate in all task suites, which is chosen as the hyperparameter for the main evaluation reported in \Cref{table:simulation-results} and \cref{table:real-results}. Interestingly, the deployment gains most with a moderate strength ($\alpha=0.6$ and $\alpha=0.4$), which further reveals that current VLAs trained with limited scale show generalist manipulation ability to a certain extent, but still gain incremental deviations during deployment. These results emphasize the importance of optimizing VLA training in a scalable path or an efficient way, and post-processing of VLA rollouts to make them more generalizable and intelligent.

\begin{figure}[t]
  \centering
  \includegraphics[width=\columnwidth]{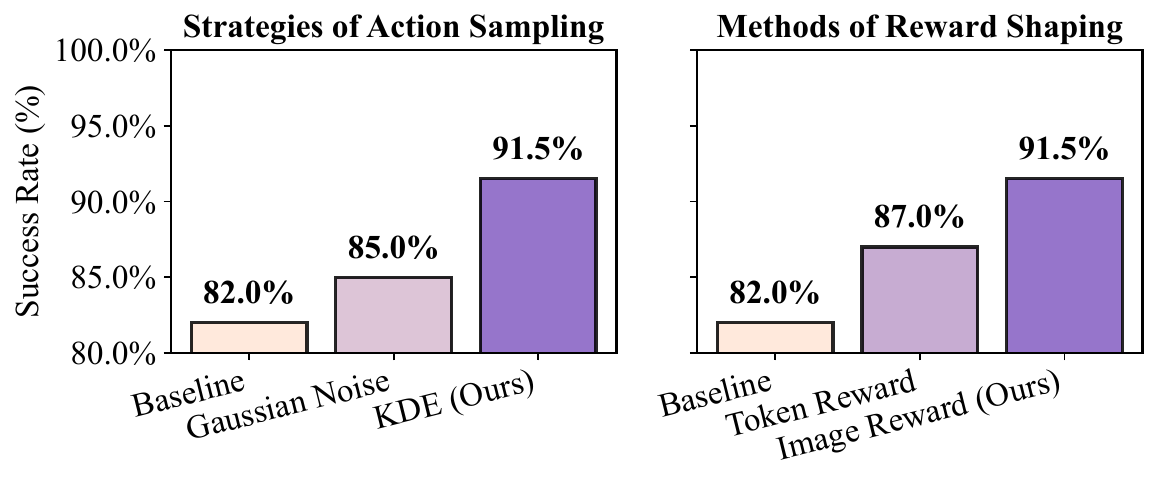}
  \caption{\small\textbf{Analysis on Techniques.} The comparison validates the design in our method, which is reflected by the significant growth in the success rate.}
  \label{fig:tech_ablation}
  \vspace{-0.3cm}
\end{figure}

\paragraph{Choices of Involved Techniques}\label{par:techniques}
As mentioned above, applying KDE on offline data enables plausible and efficient sampling, and the value estimation enables intermediate and relatively accurate state evaluation for feedback. To emphasize their applicability and advantages, we compare the KDE sampling with noisy sampling, which adds a Gaussian noise to VLA rollouts to obtain new actions with similar efficiency. The results are shown in the left part of \Cref{fig:tech_ablation}, where the KDE offers plausible choices as they implicitly reflect the smoothed expert behaviors. We compare the image-based value estimation with the token-based reward head introduced in iVideoGPT \cite{wu2024ivideogpt}. For the token-based reward head, the tokens are directly mapped to a preset value similar to \Cref{subsec:value}. The results are shown in the right part of \Cref{fig:tech_ablation}, where the image-wise value estimation can better reflect explicit task progress with a simpler implementation.

\section{Conclusion}
\label{sec:conclusion}
We identified a core limitation of current short-sighted VLA deployment and introduced \method, a plug-in framework that injects test-time reasoning into off-the-shelf VLAs, to mitigate the incremental deviations in deployment. By rolling out imagined futures with a pretrained world model and performing state-based \tech guided by a KDE action prior and an offline value estimation design, \method reflects long-horizon consequences back into the current decision. This design optimizes action selection without disturbing the execution or retraining the underlying VLA, and yields consistent gains across simulation and real-world settings with minimal additional supervision.
As the current structure still requires training before deployment to adapt to the specific task, we also see several promising directions under this framework. For example, world models with better fidelity would improve imagined state trajectories and thus the quality of feedback. Second, more principled value estimation designs, including data-driven and learning based methods, could further stabilize search. 
These facets are orthogonal to our main contribution, as \method remains a compelling paradigm for manipulation. We expect future work to build on \method and explore scalable test-time computation for general-purpose robotic manipulation.

\bibliographystyle{IEEEtran}
\bibliography{reference}

\end{document}